\documentclass[11pt,a4paper]{article}
\usepackage[hyperref]{emnlp2018}
\usepackage{times}
\usepackage{latexsym}
\usepackage{amsmath}
\usepackage{algorithm}
\usepackage{algorithmic}
\usepackage{url}
\usepackage[pdftex]{graphicx}
\usepackage{float}
\usepackage{mathrsfs}
\usepackage{amsfonts}
\usepackage{enumerate}
\usepackage{multirow}

\aclfinalcopy 
\def\aclpaperid{1177} 

\title{Jointly Multiple Events Extraction via Attention-based Graph Information Aggregation}

\author{
	Xiao Liu\textsuperscript{$\dag$} \and Zhunchen Luo\textsuperscript{$\ddagger$} \and Heyan Huang\textsuperscript{$\dag$}\thanks{\textsuperscript{*}Corresponding author.} \\
	\textsuperscript{$\dag$}School of Computer Science and Technology, Beijing Institute of Technology \\
	100081 Beijing, China \\
	{\tt \{xiaoliu,hhy63\}@bit.edu.cn} \\
	\textsuperscript{$\ddagger$}Information Research Center of Military Science, PLA Academy of Military Science \\
	100142 Beijing, China \\ 
	{\tt zhunchenluo@gmail.com}
}

\date{}

\begin{document}

\maketitle

\begin{abstract}

Event extraction is of practical utility in natural language processing.
In the real world, it is a common phenomenon that multiple events existing in the same sentence, where extracting them are more difficult than extracting a single event.
Previous works on modeling the associations between events by sequential modeling methods suffer a lot from the low efficiency in capturing very long-range dependencies.
In this paper, we propose a novel \textbf{J}ointly \textbf{M}ultiple \textbf{E}vents \textbf{E}xtraction (JMEE) framework to jointly extract multiple event triggers and arguments by introducing syntactic shortcut arcs to enhance information flow and attention-based graph convolution networks to model graph information.
The experiment results demonstrate that our proposed framework achieves competitive results compared with state-of-the-art methods.

\end{abstract}

\section{Introduction}
Extracting events from natural language text is an essential yet challenging task for natural language understanding.
When given a document, event extraction systems need to recognize event triggers with their specific types and their corresponding arguments with the roles.
Technically speaking, as defined by the ACE 2005 dataset\footnote{\url{https://catalog.ldc.upenn.edu/ldc2006t06}}, a benchmark for event extraction \cite{Grishman2005Nyu}, the event extraction task can be divided into two subtasks, i.e., event detection (identifying and classifying event triggers) and argument extraction (identifying arguments of event triggers and labeling their roles).

In event extraction, it is a common phenomenon that multiple events exist in the same sentence.
Extracting the correct multiple events from those sentences is much more difficult than in the one-event-one-sentence cases because those various types of events are often associated with each other.
For example, in the sentence ``He \textit{\textbf{left}} the company, and planned to \textit{\textbf{go}} home directly.'', the trigger word \textit{\textbf{left}} may trigger a \textit{Transport} (a person left a place) event or an \textit{End-Position} (a person retired from a company) event.
However, if we take the following event triggered by \textit{\textbf{go}} into consideration, we are more confident to judge it as a \textit{Transport} event rather than an \textit{End-Position} event.
This phenomenon is quite common in our real world, as \textit{Injure} and \textit{Die} events are more likely to co-occur with \textit{Attack} events than others, whereas \textit{Marry} and \textit{Born} events are less likely to co-occur with \textit{Attack} events.
As we investigated in ACE 2005 dataset, there are around 26.2\% (1042/3978) sentences belong to this category.


Significant efforts have been dedicated to solving this problem.
Most of them exploiting various features \cite{DBLP:conf/aaai/LiuLH016,DBLP:conf/naacl/YangM16,DBLP:conf/acl/LiJH13,DBLP:conf/emnlp/KeithHPMMO17,DBLP:conf/acl/LiuCHL016,Li2015Improving}, introducing memory vectors and matrices \cite{DBLP:conf/naacl/NguyenCG16}, introducing more transition arcs \cite{DBLP:conf/aaai/ShaQCS18}, keeping more contextual information \cite{DBLP:conf/acl/ChenXLZ015} into sentence-level sequential modeling methods like RNNs and CRFs.
Some also seek features in document-level methods \cite{DBLP:conf/acl/LiaoG10,DBLP:conf/acl/JiG08}.
However, sentence-level sequential modeling methods suffer a lot from the low efficiency in capturing very long-range dependencies while the feature-based methods require extensive human engineering, which also largely affects model performance.
Besides, these methods do not adequately model the associations between events.

\begin{figure*}[h]
	\centering
	\includegraphics[width=160mm]{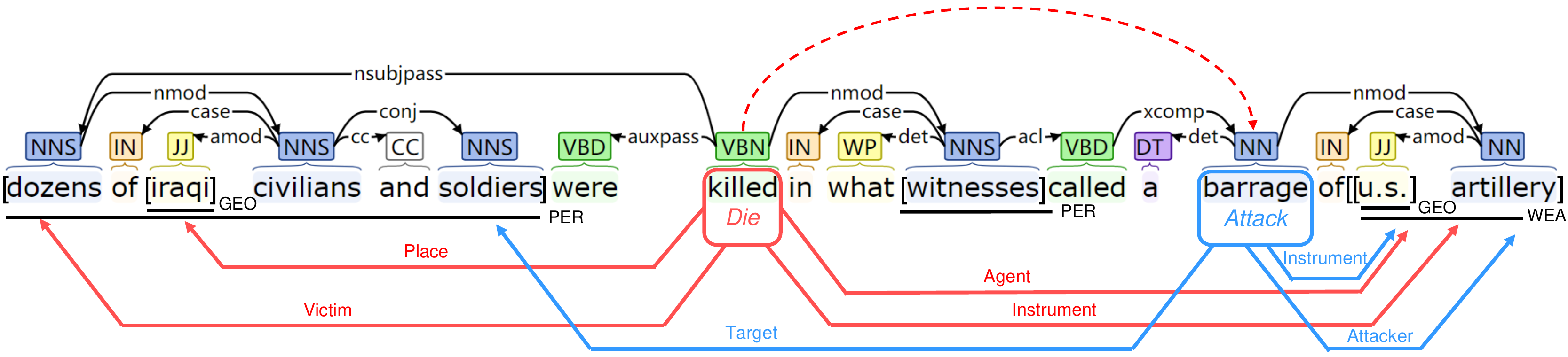}
	\vspace{-0.9cm}
	\caption{An example of dependency parsing result produced by Stanford CoreNLP. There are two events in the sentence: a \textit{Die} event triggered by the word \textit{killed} with four arguments in red and an \textit{Attack} event triggered by the word \textit{barrage} with three arguments in blue. The red dotted arc is the shortcut path consisting of three directed arcs from trigger \textit{killed} to another trigger \textit{barrage}.}
	\label{fig:1}
\end{figure*}

An intuitive way to alleviate this phenomenon is to introduce shortcut arcs represented by linguistic resources like dependency parsing trees to drain the information flow from a point to its target through fewer transitions.
Comparing to sequential order, modeling with these arcs often successfully reduce the needed hops from one event trigger to another in the same sentences.
In Figure \ref{fig:1}, for example, there are two events: a \textit{Die} event triggered by the word \textit{killed} with four arguments in red and an \textit{Attack} event triggered by the word \textit{barrage} with three arguments in blue.
We need six hops from \textit{killed} to \textit{barrage} according to sequential order, but only three hops according to the arcs in dependency parsing tree (along the \textit{nmod}-arc from \textit{killed} to \textit{witnesses}, along the \textit{acl}-arc from \textit{witnesses} to \textit{called}, and along the \textit{xcomp}-arc from \textit{called} to \textit{barrage}).
These three arcs consist of a shortcut path\footnote{In a shortcut path which consists of existing arcs, some arcs may reverse their directions.}, draining the dependency syntactic information flow from \textit{killed} to \textit{barrage} with fewer hops\footnote{The length of the longest path in a tree is always no more than the sequential length consisting of the same number of nodes, which means even in the worst cases, the shortcut path will not perform worse than sequential modeling.}.

In this paper, we propose a novel \textbf{J}ointly \textbf{M}ultiple \textbf{E}vents \textbf{E}xtraction (JMEE) framework by introducing syntactic shortcut arcs to enhance information flow and attention-based graphic convolution networks to model the graph information.
To implement modeling with the shortcut arcs, we adopt the graph convolutional networks (GCNs) \cite{DBLP:journals/corr/KipfW16,DBLP:conf/emnlp/MarcheggianiT17,DBLP:conf/aaai/NguyenG18} to learn syntactic contextual representations of each node by the representative vectors of its immediate neighbors in the graph.
And then we utilize the syntactic contextual representations to extract triggers and arguments jointly by a self-attention mechanism to aggregate information especially keeping the associations between multiple events.
Our code is released at \url{https://github.com/lx865712528/JMEE}.

We extensively evaluate the proposed JMEE framework with the widely-used ACE 2005 dataset to demonstrate its benefits in the experiments especially in capturing the associations between events.
To summary, our contribution in this work is as follows:
\begin{itemize}
    \item We propose a novel joint event extraction framework JMEE based on syntactic structures which enhance information flow and alleviate the phenomenon where multiple events are in the same sentence.
    \item We propose a self-attention mechanism to aggregate information especially keeping the associations between multiple events and prove it is useful in event extraction.
    \item We achieve the state-of-the-art performance on the widely used datasets for event extraction using the proposed model with GCNs and self-attention mechanism.
\end{itemize}

\section{Approach}

\begin{figure*}[h]
	\centering
	\includegraphics[width=160mm]{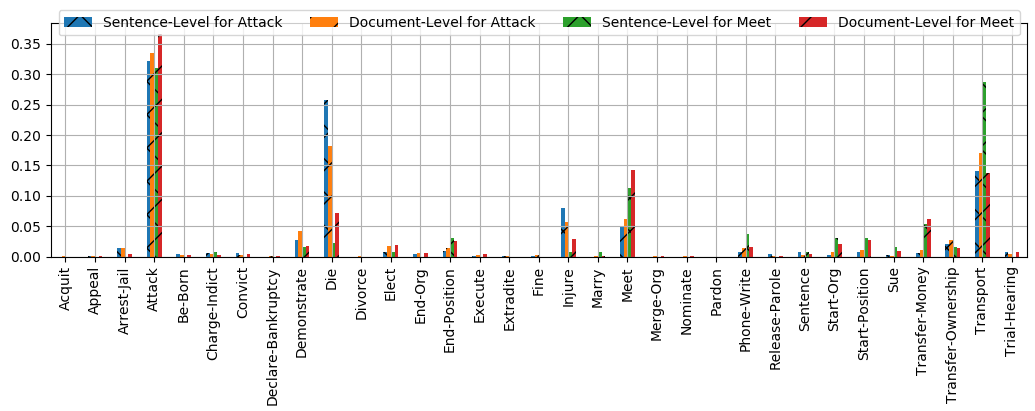}
	\vspace{-0.9cm}
	\caption{The conditional probability distributions of all the 33 event types in sentences and documents where an \textit{Attack} event or a \textit{Meet} event appears. The label on the top shows which histogram the color stands.}
	\label{fig:2}
\end{figure*}

Generally, event extraction can be cast as a multi-class classification problem deciding whether each word in the sentence forms a part of event trigger candidate and whether each entity in the sentence plays a particular role in the event triggered by the candidate triggers.
There are two main approaches to event extraction: (i) the joint approach that extracts event triggers and arguments simultaneously as a structured prediction problem, and (ii) the pipelined approach that first performs trigger prediction and then identifies arguments in separate stages.
We follow the joint approach that can effectively avoid the propagated errors in the pipeline.

Additionally, we extract events in sentence-level mainly for three reasons.
Firstly, in our investigation, we find that the document-level co-occurrence distributions of 33 types of events in the ACE 2005 dataset are relatively similar to the sentence-level co-occurrence distributions.
In Figure \ref{fig:2}, for example, the blue bars and the orange bars indicate the conditional probability distribution of all the 33 event types in sentences and documents, respectively, where an \textit{Attack} event appears.
While the green bars and the red bars indicate the sentence-level and document-level conditional probability distributions respectively.
As we can see from this figure, the top three types of \textit{Attack} event in co-occurrence relationships are \textit{Die}, \textit{Transport} and \textit{Injure}, while those of \textit{Meet} event are \textit{Attack}, \textit{Transport}, and \textit{Transfer-Money}.
Although different types of events have different co-occurrence relationships, the conditional probability distributions in two levels of the same event type are relatively similar\footnote{We only focus on the top-K co-occurrence relationships because the rest are too sparse for statistic analysis.}.
Secondly, there are many off-the-shelf sentence-level linguistic resources in the NLP community which can offer analytical information about the shortcut paths of some structures, like dependency parsing trees, AMR parsing graphs, and semantic role labeling structures.
Last but not least, we also find that events within the same sentences have more explicit relationships with each other than events in different sentences of a document, which means the associations between two events is more accessible to capture.

\begin{figure*}[h]
	\centering
	\includegraphics[width=160mm]{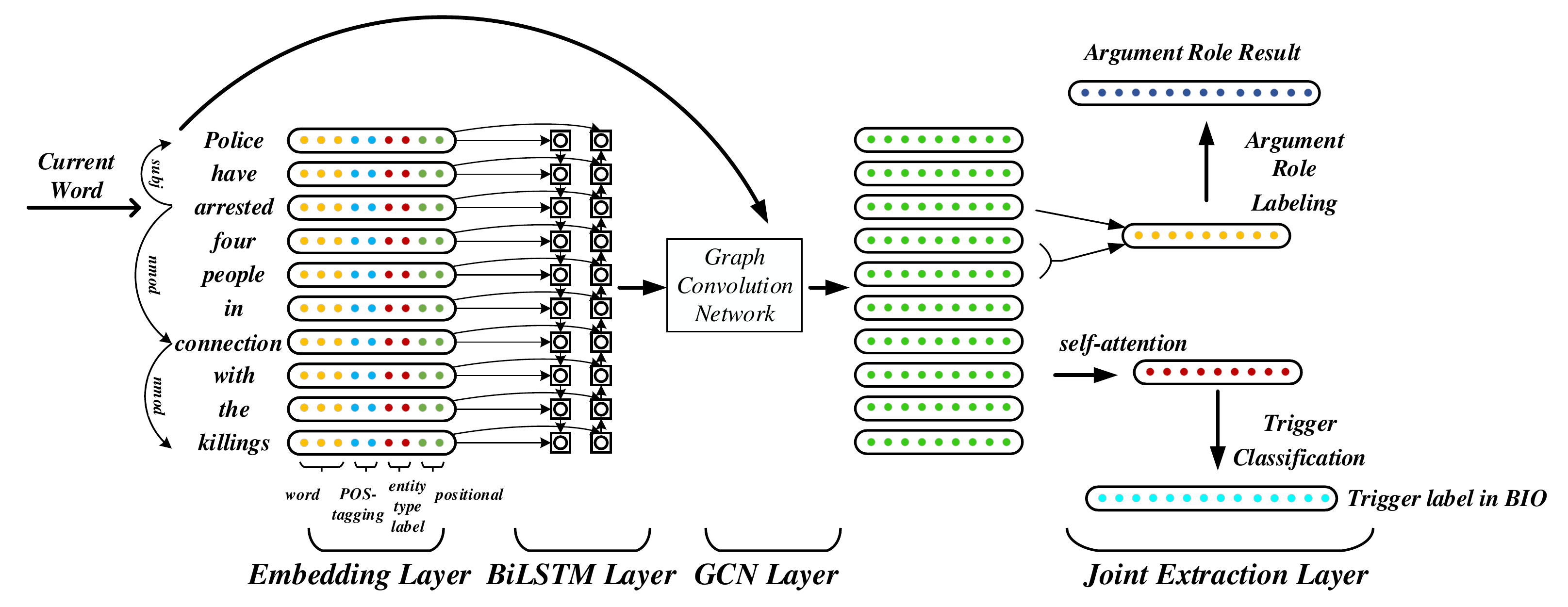}
	\vspace{-0.9cm}
	\caption{The architecture of our jointly multiple events extraction framework.}
	\label{fig:3}
\end{figure*}

Let $W=w_1,w_2,...,w_n$ be a sentence of length $n$ where $w_i$ is the $i$-th token.
Similarly, let $E=e_1,e_2,...,e_k$ be the entity mentions in the sentence where $k$ is the number of the entity mentions.
We apply the BIO annotation schema to assign trigger label $t_i$ to each token $w_i$, as there are triggers that consist of multiple tokens.
If we can get trigger candidates of certain types from trigger labels in BIO annotation schema, we then need to predict the roles (if any) that each entity mention $e_j$ plays in such events.

Our JMEE framework consists of the following four modules: (i) word representation module that represents the sentence with vectors, (ii) syntactic graph convolution network module that performs convolution operations by introducing shortcut arcs from syntactic structures, (iii) self-attention trigger classification module that captures the associations between multiple events in a sentence, and (iv) argument classification that predicts the roles each entity mention $e_j$ plays in the event candidates of specific types, as shown in Figure \ref{fig:3}.

\subsection{Word Representation}
In the word representation module, each token $w_i$ in the sentence is transformed to a real-valued vector $x_i$ by looking up in embedding matrices and concatenating the following vectors:

\begin{itemize}
    \item The word embedding vector of $w_i$: This is obtained by looking up a pre-trained word embedding matrix Glove \cite{DBLP:conf/emnlp/PenningtonSM14}.
    \item The POS-tagging label embedding vector of $w_i$: This is generated by looking up the randomly initialized POS-tagging label embedding table.
    \item The positional embedding vector of $w_i$: If $w_c$ is the current word, we encode the relative distance $i-c$ from $w_i$ to $w_c$ as a real-valued vector by looking up the randomly initialized position embedding table \cite{DBLP:conf/naacl/NguyenCG16,DBLP:conf/acl/LiuCLZ17,DBLP:conf/aaai/NguyenG18}.
    \item The entity type label embedding vector of $w_i$: Similarly to the POS-tagging label embedding vector of $w_i$, we annotate the entity mentions in a sentence using BIO annotation schema and transform the entity type labels to real-valued vectors by looking up the embedding table. It should be noticed that we use the whole entity extent in ACE 2005 dataset which contains overlapping entity mentions and we sum all the possible entity type label embedding vectors for each token.
\end{itemize}

The transformation from the token $w_i$ to the vector $x_i$ essentially converts the input sentence $W$ into a sequence of real-valued vectors $X = (x_1,x_2,...,x_n)$, which will be feed into later modules to learn more effective representations for event extraction.

\subsection{Syntactic Graph Convolution Network}
Considering an undirected graph $\mathcal{G}=(\mathcal{V}, \mathcal{E})$ as the syntactic parsing tree for sentence $W$, where $\mathcal{V}={v_1,v_2,...,v_n} (|\mathcal{V}|=n)$ and $\mathcal{E}$ are sets of nodes and edges, respectively.
In $\mathcal{V}$, each $v_i$ is the node representing token $w_i$ in $W$.
Each edge $(v_i,v_j) \in \mathcal{E}$ is a directed syntactic arc from token $w_i$ to token $w_j$, with the type label $K(w_i,w_j)$.
Additionally, to allow information to flow against the direction, we also add reversed edge $(v_j,v_i)$ with the type label $K'(w_i,w_j)$.
Following \newcite{DBLP:journals/corr/KipfW16}, we also add all the self-loops, i.e., $(v_i, v_i)$ for any $v_i \in \mathcal{V}$.
For example, in the dependency parsing tree shown in Figure \ref{fig:1}, there are four arcs in the subgraph with only two nodes \textit{``killed''} and \textit{``witnesses''}: the dependency arc with the type label $K(\textit{``killed''}, \textit{``witnesses''})=nmod$, the revresed dependency arc with the additional type label $K(\textit{``witnesses''}, \textit{``killed''})=nmod'$, and the two self-loops of \textit{``killed''} and \textit{``witnesses''} with type label $K(\textit{``killed''}, \textit{``killed''})=K(\textit{``witnesses''}, \textit{``witnesses''})=loop$.

Therefore, in the $k$-th layer of syntactic graph convolution network module,  we can calculate the graph convolution vector $h_v^{(k+1)}$ for node $v \in \mathcal{V}$ by:
\begin{equation}
  h_v^{(k+1)}=f(\sum_{u \in \mathcal{N}(v)} (W^{(k)}_{K(u,v)}h_u^{(k)}+b^{(k)}_{K(u,v)}))  
\end{equation}
where $K(u,v)$ indicates the type label of the edge $(u,v)$; $W^{(k)}_{K(u,v)}$ and $b^{(k)}_{K(u,v)}$ are the weight matrix and the bias for the certain type label $K(u,v)$, respectively; $\mathcal{N}(v)$ is the set of neighbors of $v$ including $v$ (because of the self-loop); $f$ is the activation function.
Moreover, we use the output of the word representation module $x_i$ to initialize the node representation $h_{v_i}^{0}$ of the first layer of GCNs.

After applying the above two changes, the number of predefined directed arc type label (let us say, $N$) will be doubled (to $2N+1$).
It means we will have $2N+1$ sets of parameter pairs $W^{(k)}_{k}$ and $b^{(k)}_{k}$ for a single layer of GCN.
In this work, we use Stanford Parser \cite{DBLP:conf/acl/KleinM03} to generate the arcs in dependency parsing trees for sentences as the shortcut arcs.
The current representation contains approximately 50 different grammatical relations, which is too high for the parameter number of a single layer of GCN and not compatible with the existing training data scale.
To reduce the parameter numbers, following \newcite{DBLP:conf/emnlp/MarcheggianiT17}, we modify the definition of type label $K(w_i,w_j)$ to:
\begin{equation}  
K(w_i,w_j)=
\left\{  
             \begin{array}{lr}  
             along, & (v_i,v_j) \in \mathcal{E}\\  
             rev, & i!=j \& (v_j,v_i) \in \mathcal{E}\\  
             loop, & i==j   
             \end{array}  
\right.  
\end{equation} 
where the new $K(w_i,w_j)$ only have three type labels.

As not all types of edges are equally informative for the downstream task, 
moreover, there are also noises in the generated syntactic parsing structures; we apply gates on the edges to weight their individual importances.
Inspired by \newcite{DBLP:conf/icml/DauphinFAG17,DBLP:conf/emnlp/MarcheggianiT17}, we calculate a weight $g^{(k)}_{u,v}$ for each edge $(u, v)$ indicating the importance for event extraction by:
\begin{equation}
  g^{(k)}_{u,v}=\sigma (h^{(k)}_{u}V^{(k)}_{K(u,v)}+d^{(k)}_{K(u,v)})
\end{equation}
where $\sigma$ is the logistic sigmoid function, $V^{(k)}_{K(u,v)}$ and $d^{(k)}_{K(u,v)}$ are the weight matrix and the bias of the gate.
With this additional gating mechanism, the final syntactic GCN computation is formulated as
\begin{equation}
  h_v^{(k+1)}=f(\sum_{u \in \mathcal{N}(v)} g^{(k)}_{u,v}(W^{(k)}_{K(u,v)}h_u^{(k)}+b^{(k)}_{K(u,v)}))  
\end{equation}

As stacking $k$ layers of  GCNs can model information in $k$ hops, and sometimes the length of shortcut path between two triggers is less than $k$, to avoid information over-propagating, we adapt highway units \cite{DBLP:conf/nips/SrivastavaGS15}, which allow unimpeded information flowing across stacking GCN layers.
Typically, highway layers conduct nonlinear transformation as:
\begin{equation}
    t = \sigma (W_{T}h_v^{k}+b_T)
\end{equation}
\begin{equation}
    \overline{h}_v^{(k+1)} = h_v^{(k+1)} + t \odot g(W_{H}h_v^{k}+b_H) + (1 - t) \odot h_v^{k}
\end{equation}
where $\sigma$ is the sigmoid function; $\odot$ is the element-wise product operation; $g$ is a nonlinear activation function; $t$ is called transform gate and $(1-t)$ is called carry gate.
Therefore, the input of the $k$-th GCN layers should be $\overline{h}^{(k)}$ instead of $h^{(k)}$.

The GCNs are designed to capture the dependencies between shortcut arcs, while the layer number of GCNs limits the ability to capture local graph information.
However, in this cases, we find that leveraging local sequential context will help to expand the information flow without increasing the layer number of GCNs, which means LSTMs and GCNs maybe complementary.
Therefore, instead of feeding the word representation $X = (x_1,x_2,...,x_n)$ into the first GCN layer, we follow \newcite{DBLP:conf/emnlp/MarcheggianiT17}, apply Bidirectional LSTM (Bi-LSTM) \cite{DBLP:journals/neco/HochreiterS97} to encode the the word representation $X$ as:
\begin{eqnarray}
\overrightarrow{p}_t = \overrightarrow{LSTM}(\overrightarrow{p}_{t-1},x_t) \\
\overleftarrow{p}_t = \overleftarrow{LSTM}(\overleftarrow{p}_{t-1},x_t)
\end{eqnarray}
and the input of $t$-th token to GCNs is $\overline{x}_t=[\overrightarrow{p}_t, \overleftarrow{p}_t]$, where $[,]$ is the concatenation operation.
The Bi-LSTM adaptively accumulates and abstracts the context for each token in the sentence.

\subsection{Self-Attention Trigger Classification}

When taking each token as the current word, we get the representation $D$ from all tokens calculated by GCNs.
Traditional event extraction systems often use max-pooling or its amelioration to aggregate information to each position.
However, the max-pooling aggregation mechanisms tend to produce similar results after GCN modules in our framework.
For example, if we get the aggregated vector $Ag_i$ at each position $i$ by this max-pooling mechanism $Ag_i=max\_pooling^{n}_{j=1}{(H_j)}$ with the GCNs output $\{H_j|j=1,...,n\}$ in which $n$ is the sentence length, and the vector $Ag_i$ is all the same at each position.
Besides, predicting a trigger label for a token should take other possible trigger candidates into consideration.
To capture the associations between triggers in a sentence, we design a self-attention mechanism to aggregate information especially keeping the associations between multiple events.

Given the current token $w_i$, the self-attention score vector and the context vector at position $i$ are calculated as:
\begin{equation}
    score = norm(exp(W_2f(W_1D+b_1)+b_2))
\end{equation}
\begin{equation}
    C_i=[\sum_{j=1,j!=i}^{n}score_j * D_j,D_i]
\end{equation}
where $norm$ means the normalization operation.
Then we feed the context vector $C_i$ into a fully-connected network to predict the trigger label in BIO annotation schema as:
\begin{equation}
    \overline{C}_{i} = f(W_cC_i+b_c)
\end{equation}
\begin{equation}
    y_{t_{i}} = softmax(W_t\overline{C}_{i}+b_t)
\end{equation}
where $f$ is a non-linear activation and $y_{t_{i}}$ is the final output of the $i$-th trigger label.

\subsection{Argument Classification}
When we have extracted an entire trigger candidate, which is meeting an \textit{O} label after an \textit{I-Type} label or a \textit{B-Type} label, we use the aggregated context vector $\overline{C}$ to perform argument classification on the entity list in the sentence.

For each entity-trigger pair, as both the entity and the trigger candidate are likely to be a subsequence of tokens, we aggregate the context vectors of subsequences to trigger candidate vector $T_i$ and entity vector $E_j$ by average pooling along the sequence length dimension.
Then we concatenate them together and feed into a fully-connected network to predict the argument role as:
\begin{equation}
    y_{a_{ij}} = softmax(W_a[T_i, E_j]+b_a)
\end{equation}
where $y_{a_{ij}}$ is the final output of which role the $j$-th entity plays in the event triggered by the $i$-th trigger candidate.

When training our framework, if the trigger candidate that we focus on is not a correct trigger, we set all the golden argument labels concerning the trigger candidate to \textit{OTHER} (not any roles).
With this setting, the labels of the trigger candidate will be further adjusted to reach a reasonable probability distribution.

\subsection{Biased Loss Function}
In order to train the networks, we minimize the joint negative log-likelihood loss function.
Due to the data sparsity in the ACE 2005 dataset, we adapt our joint negative log-likelihood loss function by adding a bias item as:
\begin{equation}
\begin{aligned}
    J(\theta )=-\sum_{p=1}^{N}(&\sum_{i=1}^{n_p} I(y_{t_i}) log(p(y_{t_i}|\theta)) \\
    + &\beta \sum_{i=1}^{t_p}\sum_{j=1}^{e_p}log(p(y_{a_{ij}}|\theta)))
\end{aligned}
\end{equation}
where $N$ is the number of sentences in training corpus;
$n_p$, $t_p$ and $e_p$ are the number of tokens, extracted trigger candidates and entities of the $p$-th sentence; $I(y_{t_i})$ is an indicating function, if $y_{t_i}$ is not \textit{O}, it outputs a fixed positive floating number $\alpha$ bigger than one, otherwise one; $\beta$ is also a floating number as a hyper-parameter like $\alpha$.

\section{Experiments}

\begin{table*}[h]
\centering
\begin{tabular}{|l|ccc|ccc|ccc|ccc|}
\hline
\multirow{3}{*}{\textbf{Method}} & \multicolumn{3}{c|}{\textbf{Trigger}}            & \multicolumn{3}{c|}{\textbf{Trigger}}            & \multicolumn{3}{c|}{\textbf{Argument}}           & \multicolumn{3}{c|}{\textbf{Argument}} \\
                        & \multicolumn{3}{c|}{\textbf{Identification (\%)}} & \multicolumn{3}{c|}{\textbf{Classification (\%)}} & \multicolumn{3}{c|}{\textbf{Identification (\%)}} & \multicolumn{3}{c|}{\textbf{Role (\%)}} \\ \cline{2-13} 
                        & $P$            & $R$           & $F_1$        & $P$            & $R$           & $F_1$        & $P$            & $R$           & $F_1$        & $P$        & $R$        & $F_1$     \\ \hline
Cross-Event             & \multicolumn{3}{c|}{N/A}                      & 68.7          & 68.9          & 68.8          & 50.9          & 49.7          & 50.3          & 45.1          & 44.1 & 44.6          \\ \hline
JointBeam               & 76.9          & 65.0          & 70.4          & 73.7          & 62.3          & 67.5          & 69.8          & 47.9          & 56.8          & 64.7          & 44.4 & 52.7          \\ \hline
DMCNN                   & \textbf{80.4} & 67.7          & 73.5          & 75.6          & 63.6          & 69.1          & 68.8          & 51.9          & 59.1          & 62.2          & 46.9 & 53.5          \\ \hline
PSL                     & \multicolumn{3}{c|}{N/A}                      & 75.3          & 64.4          & 69.4          & \multicolumn{3}{c|}{N/A}                      & \multicolumn{3}{c|}{N/A}             \\ \hline
JRNN                    & 68.5          & \textbf{75.7} & 71.9          & 66.0          & \textbf{73.0} & 69.3          & 61.4          & 64.2          & 62.8          & 54.2          & 56.7 & 55.4          \\ \hline
dbRNN                   & \multicolumn{3}{c|}{N/A}                      & 74.1          & 69.8          & 71.9          & 71.3          & 64.5          & 67.7          & 66.2          & 52.8 & 58.7          \\ \hline
\textbf{JMEE}                     & 80.2          & 72.1          & \textbf{75.9} & \textbf{76.3} & 71.3          & \textbf{73.7} & \textbf{71.4} & \textbf{65.6} & \textbf{68.4} & \textbf{66.8} & \textbf{54.9} & \textbf{60.3} \\ \hline
\end{tabular}
\vspace{-0.1cm}
\caption{\label{tab:1} Overall performance comparing to the state-of-the-art methods with golden-standard entities.}
\end{table*}

\subsection{Experiment Settings}
\noindent \textbf{Dataset, Resources and Evaluation Metric}

\noindent We evaluate our JMEE framework on the ACE 2005 dataset.
The ACE 2005 dataset annotate 33 event subtypes and 36 role classes, along with the NONE class and BIO annotation schema, we will classify each token into 67 categories in event detection and 37 categories in argument extraction.
To comply with previous work, we use the same data split as the previous work \cite{DBLP:conf/acl/JiG08,DBLP:conf/acl/LiaoG10,DBLP:conf/acl/LiJH13,DBLP:conf/acl/ChenXLZ015,DBLP:conf/aaai/LiuLH016,DBLP:conf/naacl/YangM16,DBLP:conf/naacl/NguyenCG16,DBLP:conf/aaai/ShaQCS18}.
This data split includes 40 newswire articles (881 sentences) for the test set, 30 other documents (1087 sentences) for the development set and 529 remaining documents (21,090 sentences) for the training set.

We deploy the Stanford CoreNLP toolkit\footnote{\url{http://stanfordnlp.github.io/CoreNLP/}} to preprocess the data, including tokenizing, sentence splitting, pos-tagging and generating dependency parsing trees.

Also, we follow the criteria of the previous work \cite{DBLP:conf/acl/JiG08,DBLP:conf/acl/LiaoG10,DBLP:conf/acl/LiJH13,DBLP:conf/acl/ChenXLZ015,DBLP:conf/aaai/LiuLH016,DBLP:conf/naacl/YangM16,DBLP:conf/naacl/NguyenCG16,DBLP:conf/aaai/ShaQCS18} to judge the correctness of the predicted event mentions.

\noindent \textbf{Hyperparameter Setting}

\noindent For all the experiments below, in the word representation module, we use 300 dimensions for the embeddings and 50 dimensions for the rest three embeddings including pos-tagging embedding, positional embedding and entity type embedding.
In the syntactic GCN module, we use a three-layer GCN, a one-layer Bi-LSTM with 220 hidden units, self-attention with 300 hidden units and 200 hidden units for the rest transformation.
We also set dropout rate to 0.5 and L2-norm to 1e-8.
The batch size in our experiments is 32, and we utilize a maximum length $n=50$ of sentences in the experiments by padding shorter sentences and cutting off longer ones. 
These hyperparameters are either randomly searched or chosen by experiences when tuning in the development set.

We use ReLU \cite{DBLP:journals/jmlr/GlorotBB11} as our non-linear activate function.
We apply the stochastic gradient descent algorithm with mini-batches and the AdaDelta update rule \cite{DBLP:journals/corr/abs-1212-5701}.
The gradients are computed using back-propagation.
During training, besides the weight matrices, we also fine-tune all the embedding tables.

\subsection{Overall Performance}

We compare our performance with the following state-of-the-art methods:

\begin{enumerate}[1]
\item \textbf{Cross-Event} is proposed by \newcite{DBLP:conf/acl/LiaoG10}, which uses document level information to improve the performance of event extraction;
\item \textbf{JointBeam} is the method proposed by \newcite{DBLP:conf/acl/LiJH13}, which extracts events based on structure prediction by manually designed features;
\item \textbf{DMCNN} is proposed by \newcite{DBLP:conf/acl/ChenXLZ015}, which uses dynamic multi-pooling to keep multiple events' information;
\item \textbf{PSL} is proposed by \newcite{DBLP:conf/aaai/LiuLH016}, which uses a probabilistic reasoning model to classify events by using latent and global information to encode the associations between events;
\item \textbf{JRNN} is proposed by \newcite{DBLP:conf/naacl/NguyenCG16}, which uses a bidirectional RNN and manually designed features to jointly extract event triggers and arguments.
\item \textbf{dbRNN} is proposed by \newcite{DBLP:conf/aaai/ShaQCS18}, which adds dependency bridges over Bi-LSTM for event extraction.
\end{enumerate}

Table \ref{tab:1} shows the overall performance comparing to the above state-of-the-art methods with golden-standard entities.
From the table, we can see that our JMEE framework achieves the best $F_1$ scores for both trigger classification and argument-related subtasks among all the compared methods.
There is a significant gain with the trigger classification and argument role labeling performances, which is 2\% higher over the best-reported models.
These results demonstrate the effectivenesses of our method to incorporate with the graph convolution and syntactic shortcut arcs.

\subsection{Effect on Extracting Multiple Events}

To evaluate the effect of our framework for alleviating the multiple events phenomenon, we divide the test data into two parts (\textbf{1/1} and \textbf{1/N}) following \newcite{DBLP:conf/naacl/NguyenCG16,DBLP:conf/acl/ChenXLZ015} and perform evaluations separately.
\textbf{1/1} means that one sentence only has one trigger or one argument plays a role in one sentence; otherwise, \textbf{1/N} is used.

\begin{table}[h]
\centering
\begin{tabular}{|l|c|c|c|c|}
\hline
\textbf{Stage}         & \textbf{Model} & \textbf{1/1}  & \textbf{1/N}  & \textbf{all}  \\ \hline
\multicolumn{1}{|c|}{} & Embedding+T    & 68.1          & 25.5          & 59.8          \\ \cline{2-5} 
                       & CNN            & 72.5          & 43.1          & 66.3          \\ \cline{2-5} 
Trigger                & DMCNN          & 74.3          & 50.9          & 69.1          \\ \cline{2-5} 
                       & JRNN           & \textbf{75.6} & 64.8          & 69.3          \\ \cline{2-5} 
                       & \textbf{JMEE}           & 75.2          & \textbf{72.7} & \textbf{73.7} \\ \hline
                       & Embedding+T    & 37.4          & 15.5          & 32.6          \\ \cline{2-5} 
                       & CNN            & 51.6          & 36.6          & 48.9          \\ \cline{2-5} 
Argument               & DMCNN          & 54.6          & 48.7          & 53.5          \\ \cline{2-5} 
                       & JRNN           & 50.0          & 55.2          & 55.4          \\ \cline{2-5} 
                       & \textbf{JMEE}           & \textbf{59.3} & \textbf{57.6} & \textbf{60.3} \\ \hline
\end{tabular}
\vspace{-0.3cm}
\caption{\label{tab:2} System Performance on Single Event Sentences \textbf{(1/1)} and Multiple Event Sentences \textbf{(1/N)}}
\end{table}

Table \ref{tab:2} illustrates the performance ($F_1$ scores) of \textbf{JRNN} \cite{DBLP:conf/naacl/NguyenCG16}, \textbf{DMCNN} \cite{DBLP:conf/acl/ChenXLZ015}, the two baseline model \textbf{Embedding+T} and \textbf{CNN} in \newcite{DBLP:conf/acl/ChenXLZ015} and our framework in trigger classification subtask and argument role labeling subatsk.
\textbf{Embedding+T} uses word embedding vectors and the traditional sentence-level features in \newcite{DBLP:conf/acl/LiJH13}, while \textbf{CNN} is similar to \textbf{DMCNN}, except that it applies the standard max-pooling mechanism instead of the dynamic multi-pooling mechanism.
We can see that our framework significantly outperforms all the other methods, especially in trigger classification subtask.
In the \textbf{1/N} data split of triggers, our framework is 7.9\% better than the \textbf{JRNN}, which demonstrates that our method of leveraging syntactic shortcut arcs and self-attention aggregation mechanism is helpful in alleviating the multiple events phenomenon.

\subsection{Analysis of Self-Attention Mechanism}

\begin{figure}[h]
	\centering
	\includegraphics[width=86mm]{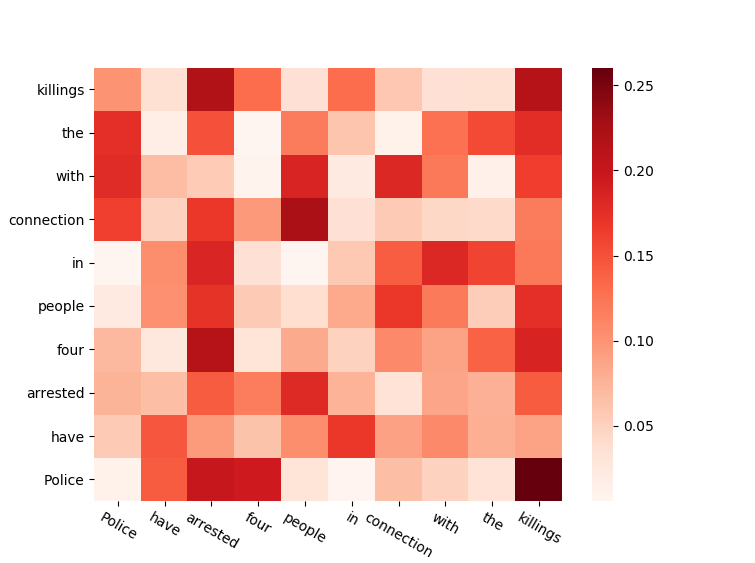}
	\vspace{-0.9cm}
	\caption{Visualization of the attention scores of a sentence containing two events: an \textit{Arrest-Jail} event triggered by \textit{arrested} and a \textit{Die} event triggered by \textit{killings}. Each row is the group of scores derived by the self-attention aggregation mechanism. Darker red means higher score and stronger interaction.}
	\label{fig:4}
\end{figure}

We use a sentence ``police have arrested four people in connection with the killing'' as an example to illustrate the captures features in our self-attention aggregation mechanism by transforming the attention scores to a row-wise heap map in Figure \ref{fig:4}.
There are two events in the sentence: an \textit{Arrest-Jail} event triggered by \textit{arrested} and a \textit{Die} event triggered by \textit{killings}.
Additionally, the entity \textit{police} plays an \textit{Agent} role and the entity \textit{four people} plays a \textit{Person} role in the \textit{Arrest-Jail} event.

As we can see from the Figure \ref{fig:4}, in the row of \textit{arrested}, there are relatively strong connections with \textit{arrested} (self), \textit{four people} (its argument) and \textit{killings} (other event).
And in the row of \textit{killings}, there are also relatively strong connections with \textit{killings} (self) and \textit{arrested} (other event).
Besides, the words \textit{police}, \textit{four} and \textit{in} also have high scores with \textit{killings}, which may mean be on account of the context information propagation though syntactic shortcut arcs.

\section{Related Work}

There are several existing approaches exploiting the associations between events in the event extraction task.
Some of them alleviate this phenomenon by exploiting various sentence-level features, such as ranking dependencies \cite{DBLP:conf/acl/McCloskySM11}, combinational features of triggers and arguments \cite{DBLP:conf/acl/LiJH13}, probabilistic soft logic information \cite{DBLP:conf/aaai/LiuLH016,DBLP:conf/acl/LiuCHL016}, trigger-specific features and relational features \cite{DBLP:conf/naacl/YangM16,DBLP:conf/emnlp/KeithHPMMO17}.
Others also seek features in document-level methods \cite{DBLP:conf/acl/LiaoG10,DBLP:conf/acl/JiG08,DBLP:conf/acl/HongZMYZZ11,DBLP:conf/naacl/ReichartB12,DBLP:conf/acl/LuR12}.
The feature-based methods require extensive human engineering, which also essentially affects model performances, and learn them from the unbalanced training data, however, it is difficult for sparse events.

There are also a group of deep learning methods using RNNs \cite{DBLP:conf/naacl/NguyenCG16,DBLP:conf/aaai/ShaQCS18,DBLP:conf/aaai/Liu00018} and CNNs \cite{DBLP:conf/acl/ChenXLZ015,DBLP:conf/acl/FengHTJQL16,DBLP:conf/emnlp/NguyenG16} capturing the associations between events.
However, sentence-level sequential modeling methods suffer a lot from the low efficiency in capturing very long-range dependencies.
Besides, these methods do not fully model the associations between events.

\section{Conclusion and Future Work}
This paper presents a novel deep neural jointly multiple events extraction (JMEE) framework for the task of event extraction, especially for alleviating the multiple-event phenomenon.
In our framework, we introduce syntactic shortcut arcs to enhance information flow and adapt the graph convolution network to capture the enhanced representation.
Then a self-attention aggregation mechanism is applied to aggregate the associations between events.
Besides, we jointly extract event triggers and arguments by optimizing a biased loss function due to the imbalances in the dataset.
The experiment results demonstrate the effectiveness of our proposed framework.
In the future, we plan to exploit the information of one argument which plays different roles in various events to do better in event extraction task.

\section*{Acknowledgments}

We would like to thank Yansong Feng, Ying Zeng, Xiaochi Wei, Qian Liu and Changsen Yuan for their insightful comments and suggestions.
We also very appreciate the comments from anonymous reviewers which will help further improve our work.
This work is supported by National Natural Science Foundation of China (No. 61751201 and No. 61602490) and National Key R\&D Plan (No. 2017YFB0803302).

\bibliography{emnlp2018}
\bibliographystyle{acl_natbib_nourl}

\end{document}